\documentclass[runningheads]{llncs}
\usepackage{graphicx}
\usepackage{amsmath,amssymb} 
\usepackage{color}
\usepackage{wrapfig}
\usepackage{multirow}

\begin{document}

\title{DOCK: Detecting Objects \\
	by transferring Common-sense Knowledge}
\titlerunning{DOCK: Detecting Objects by transferring Common-sense Knowledge}

\author{Krishna Kumar Singh\inst{1,3}\orcidID{0000-0002-8066-6835} \and \\
	Santosh Divvala\inst{2,3}\orcidID{0000-0003-4042-5874} \and
	Ali Farhadi\inst{2,3}\orcidID{0000-0001-7249-2380} \and
	Yong Jae Lee\inst{1}\orcidID{0000-0001-9863-1270}}

\institute{$^{1}$University of California, Davis $^{2}$University of Washington $^{3}$Allen Institute for AI
\url{https://dock-project.github.io}}

\authorrunning{K. K. Singh, S. Divvala, A. Farhadi, and Y. J. Lee}

\maketitle
\vspace{-0.1in}
\begin{abstract}
We present a scalable approach for Detecting Objects by transferring Common-sense Knowledge (DOCK) from source to target categories. In our setting, the training data for the source categories have bounding box annotations, while those for the target categories only have image-level annotations.  Current state-of-the-art approaches focus on image-level visual or semantic similarity to adapt a detector trained on the source categories to the new target categories.  In contrast, our key idea is to (i) use similarity not at the image-level, but rather at the region-level, and (ii) leverage richer common-sense (based on attribute, spatial, etc.) to guide the algorithm towards learning the correct detections. We acquire such common-sense cues automatically from readily-available knowledge bases without any extra human effort. On the challenging MS COCO dataset, we find that common-sense knowledge can substantially improve detection performance over existing transfer-learning baselines.

\end{abstract}

\section{Introduction}

Object detection has witnessed phenomenal progress in recent years, where {\em fully-supervised} detectors have produced amazing results. However, getting large volumes of bounding box annotations has become an Achilles heel of this setting. To address this scalability concern, transfer-learning methods that transform knowledge from source categories (with bounding boxes) to {\em similar} target classes (with only image labels) have evolved as a promising alternative~\cite{shi-bmvc2012,hoffman-nips2014,rochan-cvpr2015,tang-cvpr2016}.

\begin{wrapfigure}{r}{0.3\textwidth}
  \vspace{-0.4in}
  \begin{center}
    \includegraphics[width=0.275\textwidth]{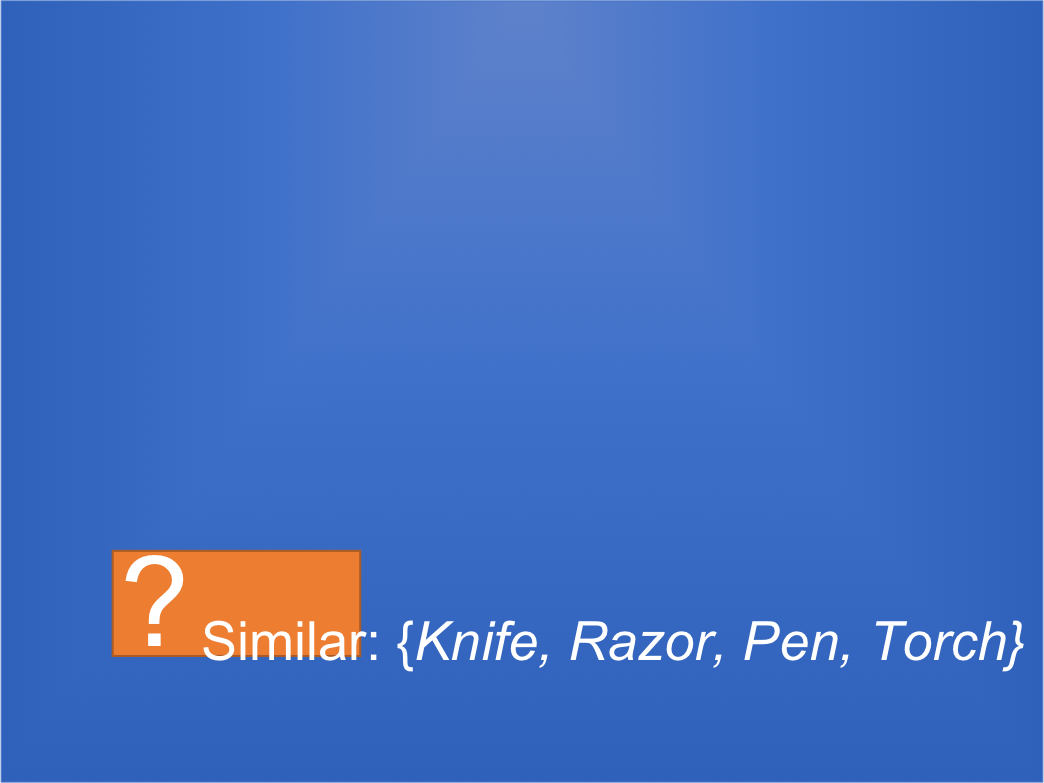}
  \end{center}
  \vspace{-0.3in}
  \caption{\small{Guess the object?}}
  \vspace{-0.2in}
  \label{fig:teaser1a}
\end{wrapfigure}

While recent works~\cite{hoffman-nips2014,hoffman-cvpr2015,tang-cvpr2016} have demonstrated the exciting potential of transfer-learning on {\em object-centric} datasets like ImageNet, it has not yet been thoroughly explored on more complex {\em scene-centric} datasets like MS COCO. Why is it so?

We hypothesize three key challenges:

(i) Existing transfer learning methods rely only on similarity knowledge between the source and target categories to compute the transformations that need to be transferred. Unfortunately, using similarity alone is often insufficient. For example, can you guess the orange-colored region proposal in the masked image using only the provided similarity cues (Fig.~\ref{fig:teaser1a})?  (ii) Existing methods depend on having a robust image-level object classifier for transferring knowledge, which for object-centric datasets like ImageNet is easy to obtain, but is challenging for scene-centric datasets like MS COCO (where multiple and potentially small objects like `toothbrush' exist in each image). If the image classifier does not perform well, then transforming it into a detector will not perform well either. (iii) Finally, if instances of the target classes frequently co-occur with the source classes, then the target class regions can end up being undesirably learned as `background' while training the detector for the source classes.

\begin{wrapfigure}{R}{0.35\textwidth}
  \vspace{-0.5in}
  \begin{center}
    \includegraphics[width=0.34\textwidth]{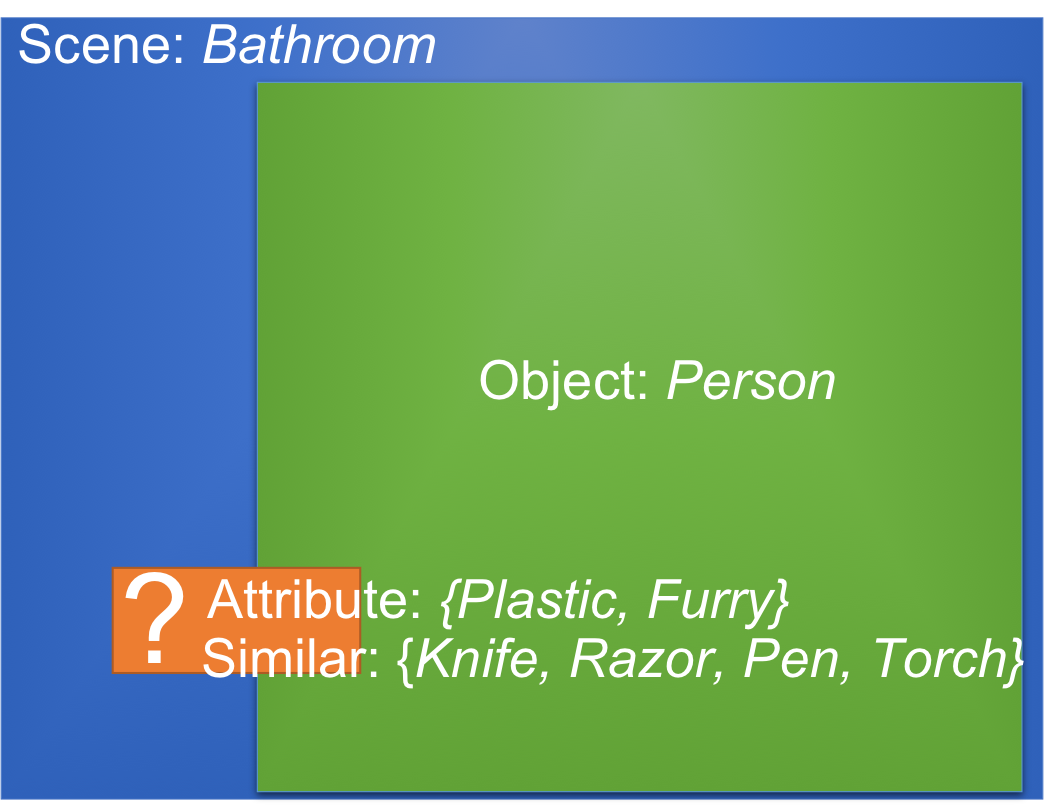}
  \end{center}
    \vspace{-0.3in}
  \caption{\small{Using the multiple common-sense cues, can you now guess the object corresponding to the orange box?  (For answer, see~\cite{teaser-answer})}}
    \vspace{-0.3in}
  \label{fig:teaser1}
\end{wrapfigure}

In this paper, we overcome the above limitations by proposing a new approach for Detecting Objects by transferring Common-sense Knowledge (DOCK). To overcome the first limitation, our key idea is to leverage multiple sources of \emph{common-sense} knowledge.  Specifically, we encode: (1) \emph{similarity}, (2) \emph{spatial}, (3) \emph{attribute}, and (4) \emph{scene}. For example, if `spoon' is one of the target objects and `fork', `table', and `kitchen' are among the source categories, we can learn to better detect the `spoon' by leveraging the fact that it is usually `{\em similar} to fork', `{\em on} a table', `{\em is} metallic', and `{\em seen} in kitchens'. Fig.~\ref{fig:teaser1} shows another scenario, which builds upon the Fig.~\ref{fig:teaser1a} example.

In this way, even if a target class does not have a visually/semantically {\em similar} class among the source classes, the other common-sense can help in obtaining a better detector. All the common-sense knowledge we use is freely-acquired from readily-available external knowledge bases~\cite{liu-bt2004,navigli-ai2012,miller-acm1995,tandon-wsdm2014,krishna-ijcv2017,yatskar-cvpr2016}. Further, our approach learns all the required common-sense models using source-class bounding box annotations only and does not require \emph{any} bounding box annotations for the target categories. 

To address the latter limitations, our idea is to directly model objects at the region-level rather than at the image-level.  To this end, any detection framework that learns using region proposals with image-level labels for the target object categories is applicable. In this paper, we use an object proposal classification and ranking detection framework based upon~\cite{bilen-cvpr2016} for its simplicity and competitive performance. It learns an object detector by minimizing an image classification loss based on object proposals' class probabilities. We inject common-sense into this framework by modulating the object proposals' class probabilities with our proposed common-sense prior probabilities. The proposed priors give higher preference to regions that are more likely (under common-sense) to belong to the object-of-interest.  Interestingly, since the common-sense cues are encoded only as \emph{priors}, our algorithm can choose to ignore them when they are not applicable. This is particularly helpful to alleviate the concern when frequently co-occurring target classes  are incorrectly learned as `background'.

We evaluate our approach on the challenging MS COCO dataset~\cite{lin-eccv2014}.  We find that transferring common-sense knowledge substantially improves object detection performance for the target classes that lack bounding box annotations, compared to other contemporary transfer-learning methods~\cite{hoffman-nips2014,hoffman-cvpr2015,tang-cvpr2016}.  We also perform ablation analyses to inspect the contribution of our proposed idea of encoding common sense.  Finally, we explore the potential of our proposed framework in the context of webly-supervised object detection.

\section{Related Work}

\paragraph{Transfer learning for scalable object detection.} 
Existing transfer learning approaches can be roughly divided into two groups: one that learns using bounding box annotations for both source and target categories~\cite{salakhutdinov-cvpr2011,aytar-iccv2011,lim-nips2011,donahue-cvpr2013,xu-pami2014,wang-cvpr2015}, and another that learns using bounding box annotations for source categories but only image-level annotations for target categories~\cite{shi-bmvc2012,hoffman-nips2014,hoffman-cvpr2015,rochan-cvpr2015,tang-cvpr2016}.  In this paper, we are interested in the latter setting, which is harder but likely to be more scalable.  In particular, the recent state-of-the-art deep learning approaches of~\cite{hoffman-nips2014,hoffman-cvpr2015,tang-cvpr2016} adapt an image classifier into an object detector by learning a feature transformation between classifiers and detectors on the source classes, and transfer that transformation to related target classes based on visual or semantic similarity for which only classifiers exist. While our approach also leverages pre-trained object detectors for encoding visual and semantic similarity, we explore additional common-sense cues such as spatial and attribute knowledge. Moreover, both~\cite{hoffman-nips2014,tang-cvpr2016} use similarity information at the image-level (which works well on only object-centric datasets like ImageNet), while our approach uses similarity at the region-level. All these contributions together help us achieve a significant performance boost on the scene-centric MS COCO dataset.

\paragraph{Use of context.} Our use of common-sense is related to previous works on context which leverage additional information beyond an object's visual appearance. Context has been used for various vision tasks including object detection~\cite{rabinovich-iccv2007,divvala-cvpr2009,desai-ijcv2011,russakovsky-eccv2012,bilen-cvpr2014,gidaris-iccv2015}, semantic segmentation~\cite{mottaghi-cvpr2014}, and object discovery~\cite{lee-tpami2012,doersch-eccv2014}. As context by definition is something that frequently co-occurs with the object-of-interest, without bounding box annotations the contextual regions can easily be confused with the object-of-interest (e.g., a piece of `road' context with a `car').  Our approach tries to address this issue by using external common-sense knowledge to model context for a target object in terms of its spatial relationship with previously-learned source objects. This idea is related to~\cite{lee-tpami2012}, which makes use of already known objects to discover new categories from unlabeled images.

\paragraph{Using external knowledge for vision tasks.} Our field has witnessed the rise of several interesting knowledge bases, including ConceptNet~\cite{liu-bt2004}, BabelNet~\cite{navigli-ai2012}, WordNet~\cite{miller-acm1995}, WebChild~\cite{tandon-wsdm2014}, Visual Genome~\cite{krishna-ijcv2017}, ImSitu~\cite{yatskar-cvpr2016}, etc. While resources like WebChild~\cite{tandon-wsdm2014} and BabelNet~\cite{navigli-ai2012} are created automatically by crawling the web, others are generated with crowd-sourced effort.  The key advantage of these resources is that they contain freely-available rich knowledge.

Such external knowledge bases have been used in several vision tasks including image classification~\cite{marino-cvpr2017,tandon-wsdm2018}, VQA~\cite{zhu-arxiv2015,wu-cvpr2016}, visual relationship detection~\cite{lu-eccv2016,plummer-iccv2017}, and modeling object affordances~\cite{zhu-eccv2014}. However, there has been very limited work on using external knowledge for object detection~\cite{fang-ijcai2017}, especially in the transfer learning setting in which bounding box annotations for the target classes are lacking. Tang et al.~\cite{tang-cvpr2016} use word2vec semantic similarity between classes to perform domain transfer between a classifier and a detector. In contrast, we go beyond using semantic similarity and explore spatial, scene, and attribute cues.

\section{Proposed Approach}

In this section, we first briefly describe the base detection network used in our framework and then explain our proposed approach for injecting common-sense knowledge into it. Finally, we describe our process for automatically gathering the different types of common-sense knowledge from existing resources.

\subsection{Base Detection Network}
\label{sec:baseline}

Our idea of transferring common-sense knowledge to improve object detection is generic and could be incorporated into any detection approach that learns from image-level labels. In our work, we use an object proposal classification and ranking framework based on~\cite{bilen-cvpr2016} for its simplicity and end-to-end nature.

\begin{figure*}[t!]
    \centering
   \includegraphics[width=1\textwidth]{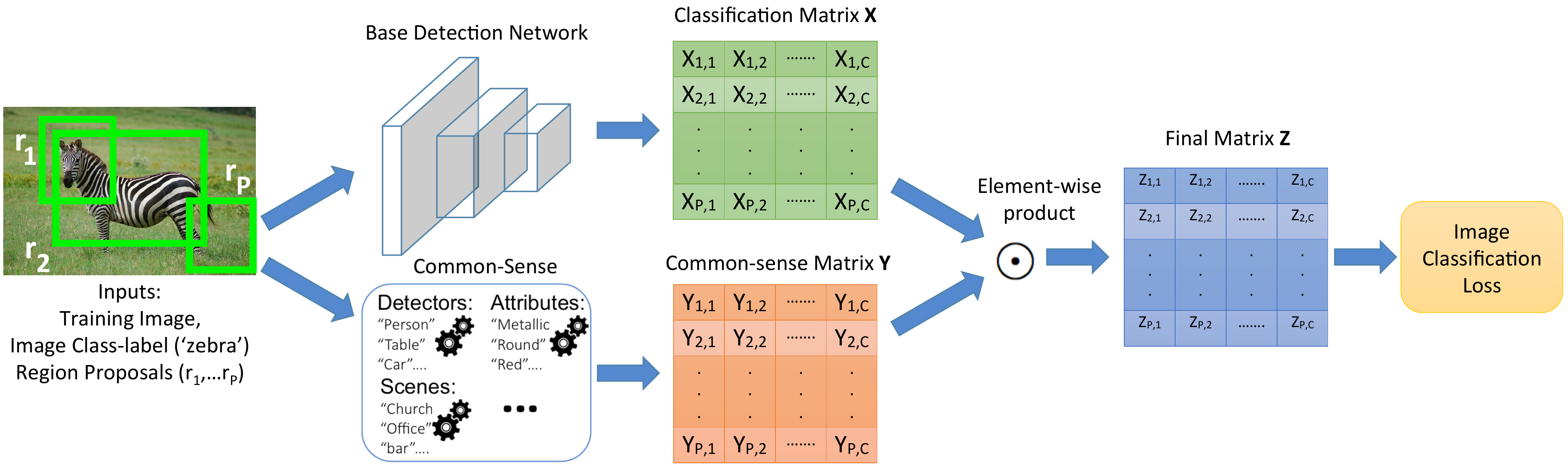}
   \vspace{-0.2in}
    \caption{Proposed framework for transferring common-sense knowledge for object detection. The base detection network computes a classification matrix $X_{P \times C}$ without any bounding box annotations (Section~\ref{sec:baseline}). We introduce a common-sense matrix $Y_{P \times C}$ that modulates the probabilities of region proposals belonging to various classes based on common-sense knowledge \label{sec:baseline} (Section~\ref{sec:useCS}). The common-sense matrix is computed using readily-available knowledge base resources (Section~\ref{sec:getCS}).}
    \label{fig:approach}
\end{figure*}

The initial layers of the network consist of convolution layers followed by spatial pyramid pooling layers to pool features corresponding to image region proposals ($r_i$). After pooling, the network has two data streams: the {\em recognition} stream assigns a classification score for each region proposal by applying a softmax over the classes to produce a $P \times C$ recognition matrix $X_{r}$, whereas the {\em detection} stream assigns probability of a region proposal to be selected for a specific class by applying a softmax over the proposals to produce a $P \times C$ detection matrix $X_{d}$. The final probability for each proposal to belong to different classes is computed by taking their element-wise dot product $X = X_{r} \odot X_{d}$.
The network takes $P$ proposals of a training image as input and outputs the probability for each of them to belong to $C$ classes. This is shown as a $P \times C$ classification matrix $X$ in Fig.~\ref{fig:approach}. Note that the network learns to detect objects while being trained for the image classification task. The image-level class probabilities are obtained by summing the probabilities of each class ($c_i$) over the proposals:
$$
Prob(c_i) = \sum_{n=1}^{P} X_{r_n,c_i} \space, i \in (1,C),
$$
where $X_{r_n,c_i}$ is the probability of proposal $r_n$ belonging to class $c_i$. A binary cross-entropy loss is applied over the probabilities to learn the detection models.

\subsection{Transferring Common-sense}
\label{sec:useCS}

In order to transfer common-sense knowledge from the source categories with both image and bounding box annotations to the target categories with only image-level annotations, we augment the above base detection network with a novel {\em common-sense} matrix $Y$ of size $P \times C$ (analogous to the classification matrix $X_{P \times C}$). Each element of $Y_{r_n,c_i}$ can be thought of as representing a `prior' probability of a proposal $r_n$ belonging to class $c_i$ according to common-sense knowledge (see Fig.~\ref{fig:approach}). We will maintain a separate common-sense matrix for each type of common-sense ({\em similarity}, {\em attribute}, etc.) and later (Section~\ref{sec:getCS}) describe the details for acquiring and merging these matrices.

Assuming we have access to this common-sense matrix $Y$, we utilize this information by taking an element-wise dot product of it with the classification matrix ($X$) to create a resultant matrix $Z_{P \times C}$:
$$ Prob(c_i) = \sum_{n=1}^{P} Y_{r_n,c_i} \ast X_{r_n,c_i} = \sum_{n=1}^{P} Z_{r_n,c_i} \space, i \in (1,C),$$
which now will be used for obtaining the image-level class probabilities over which a binary cross-entropy loss is applied.

For example, in Fig.~\ref{fig:approach}, the {\em attribute} common-sense matrix (which would encode the common-sense that a `zebra' is {\em striped}) would have a low prior probability ($Y_{r_P,zebra}$) for the proposal $r_P$ to be a `zebra'. And, the {\em class-similarity} common-sense (which would encode the common-sense that a `zebra' is similar to a `horse') would have a low value for the zebra head proposal ($Y_{r_1,zebra}$) compared to the zebra full-body proposal ($Y_{r_2,zebra}$).

Intuitively, the common-sense matrix $Y$ influences the values of the classification matrix $X$ during training and over time the common-sense priorities are transferred from $Y$ to $X$. To drive home this intuition, we take the example in Fig.~\ref{fig:approach}. The $Prob_{zebra}$ should be high for this training image as it contains a `zebra', i.e.,
$\sum_{n=1}^{P} Z_{r_n,zebra}= \sum_{n=1}^{P} Y_{r_n,zebra} \ast X_{r_n,zebra}$ should be high. This can be easily achieved if both $Y_{r_n,zebra}$ and $X_{r_n,zebra}$ are high. In this case $Y_{r_2,zebra}$ is high according to common-sense which in turn encourages the network to have high value for $X_{r_2,zebra}$. At the same time, due to a low value for $Y_{r_1,zebra}$ and $Y_{r_P,zebra}$, the network is discouraged to have a high value for $X_{r_1,zebra}$ and $X_{r_P,zebra}$. Therefore, during training itself, the network learns to incorporate the common-sense prior information of the $Y$ matrix into the $X$ matrix.

\subsection{Acquiring Common-sense}
\label{sec:getCS}

Now that we have seen how we transfer common-sense information (i.e., matrix $Y$) into the base detection framework, we next explain our approach for automatically gathering this matrix using existing knowledge resources.

\begin{figure}[t]
    \centering
    \includegraphics[width=0.8\textwidth]{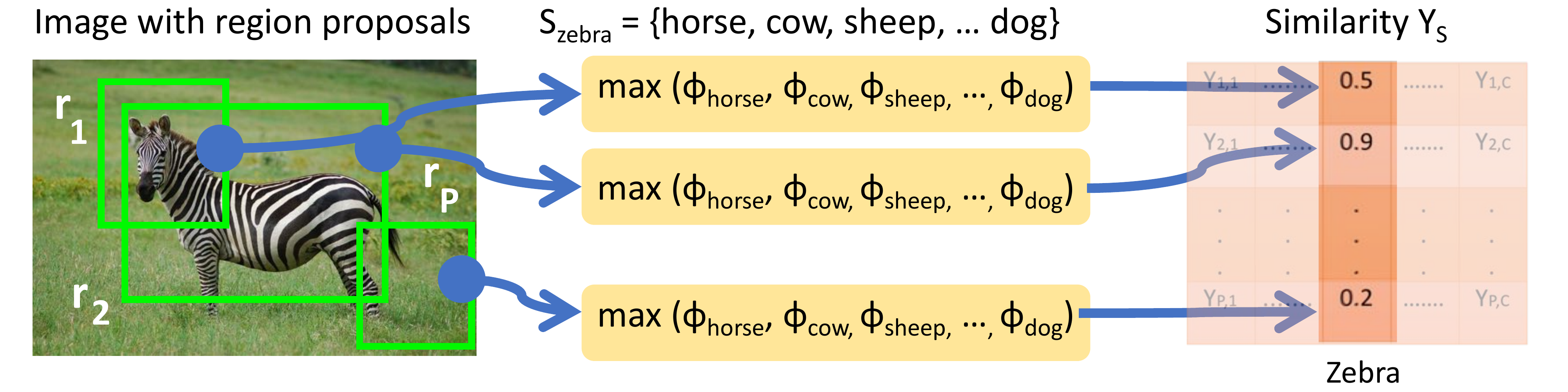}
    \vspace{-0.2in}
    \caption{\textbf{Similarity Common-sense:} For computing the $Y_s(.,zebra)$ values, all the proposals of the input image are scored by the detectors of zebra's semantically similar classes $S_{zebra}$. Observe that proposal $r_2$, which contains the zebra's full-body gets the highest $Y_s$ value.}
    \label{fig:commonsense_sim}
\end{figure}

\paragraph{Class Similarity Common-sense.} Our goal here is to leverage the semantic similarity of a new target class to previously-learned source classes. For example, as a `zebra' is semantically similar to a `horse', the proposals that are scored higher by a `horse' detector should be more probable of being a `zebra'. More generally, for any class $c_i$, the proposals looking similar to its semantically similar classes should receive higher prior for $c_i$.

To construct the class-similarity common-sense matrix $Y_s$, we tap into the readily-available set of pre-trained detectors ($\phi$) for the source object classes ($C_{source}$) in the PASCAL VOC~\cite{everingham-voc2012} knowledge base. Let $c_i$ be one of the new target object classes for which we are trying to learn a detector with just image-level labels. To find the set of semantically similar source classes (i.e., $S_{c_i} \subset C_{source}$) to $c_i$, we represent all the classes ($c_i$ as well as $C_{source}$) using their word2vec textual feature representation~\cite{mikolov-nips2013} and then compute the cosine-similarity between them. We choose all classes from $C_{source}$ with cosine-similarity above a threshold ($0.35$) as $S_{c_i}$.

We use the detectors of the classes in $S_{c_i}$ to compute the values in $Y_s$ as:

$$ Y_s(r_n,c_i) = \max_{c_j \in S_{c_i} }\phi_{c_j}(r_n), n\in(1,P).$$
Specifically, we set the value $Y_s(r_n,c_i)$ for a proposal $r_n$ to be of class $c_i$ as being equal to the maximum detection probability of the classes similar to $c_i$. Fig.~\ref{fig:commonsense_sim} illustrates how the class-similarity common-sense $Y_s$ is assigned in case of the target `zebra' object class, where $S_{zebra}$ consists of the source object classes \{`horse',`cow',`sheep',`dog',`cat',`bird'\}. Observe that a correct proposal (i.e., $r_2$ containing the `zebra' full-body) gets the highest similarity common-sense probability as it is scored higher by the object detectors in $S_{zebra}$.

\paragraph{Attribute Common-sense.} Attributes are mid-level semantic visual concepts (e.g., {\em furry}, {\em red}, {\em round}, etc.,) that are shareable across object categories~\cite{farhadi-cvpr2009}. For example, an `apple' is usually {\em red}, a `clock' is typically {\em round}, etc. Therefore region proposals that possess the characteristic attributes of a specific class should be more probable to belong to that class.

To build the attribute common-sense matrix $Y_a$, we leverage the pre-trained set of attribute classifiers ($\theta$) from the ImageNet Attribute~\cite{russakovsky-eccv2010} knowledge base and the readily-available set of object-attribute relationships from the Visual Genome~\cite{krishna-ijcv2017} knowledge base. Let $c_i$ be one of the new target classes for which we are trying to learn a detector and $A_{c_i}$ be its set of common attributes (determined by the frequency by which it is used to described  $c_i$). The classifiers of the attributes in $A_{c_i}$ are used to compute the values of the matrix $Y_a$ as
$$ Y_a(r_n,c_i) = \max_{a_j \in A_{c_i} }\theta_{a_j}(r_n), n\in(1,P).$$

\begin{figure}[t]
    \centering
    \includegraphics[width=0.8\textwidth]{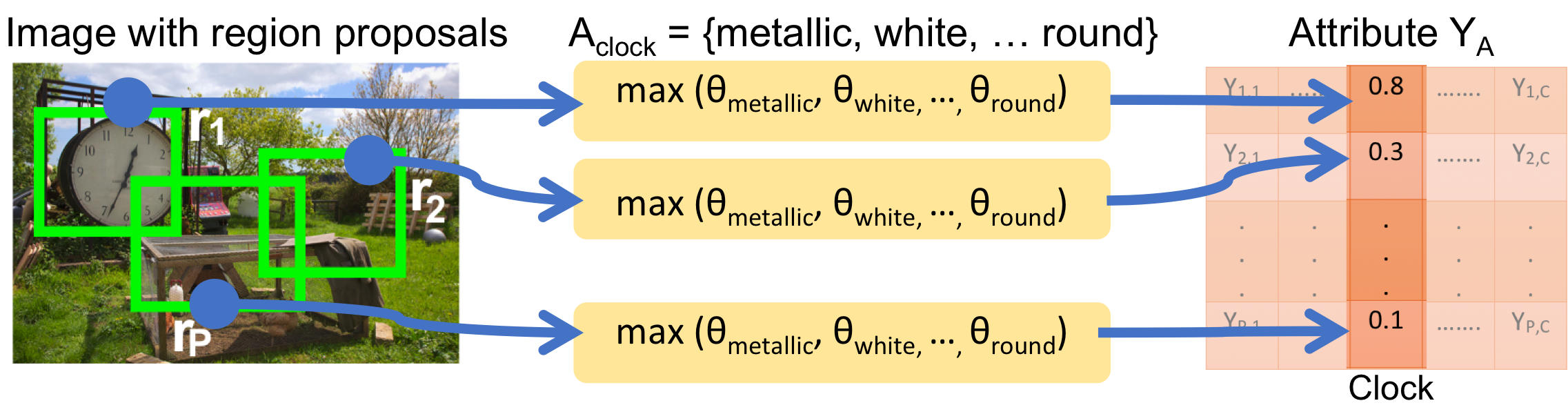}
    \vspace{-0.2in}
    \caption{\textbf{Attribute Common-sense:} For computing $Y_a(.,clock)$ values, we apply the attribute classifiers for the common attributes of `clock' on the proposals. Observe that $r_1$ containing a `clock' with {\em white} \& {\em round} attributes gets the highest $Y_a$ value.}
    \label{fig:commonsense_att}
\end{figure}

As the attributes in ImageNet~\cite{russakovsky-eccv2010} knowledge base (that we use in this work) have been additionally grouped into sets of {\em color} $A^{col}$, {\em shape} $A^{shape}$, and {\em texture} $A^{text}$ attributes, we adopt this information by updating $Y_a(r_n,c_i)$ as: $ Y_a(r_n,c_i) = mean(Y^{col}_a(r_n,c_i), Y^{shape}_a(r_n,c_i), Y^{text}_a(r_n,c_i)),$ where $Y^{col}$, $Y^{shape}$, $Y^{text}$ have been computed over the $A^{col}$, $A^{shape}$, $A^{text}$ domains. 
In Fig.~\ref{fig:commonsense_att}, for the `clock' class, the proposal $r_1$ containing its attributes, i.e., {\em white/black} and {\em round}, get the highest $Y$ value, while the other proposals ($r_2$, $r_P$) get lower values as they lack the characteristic `clock' attributes.

\paragraph{Spatial Common-sense.} In our day-to-day experience objects often appear in characteristic spatial relations with other objects. For example, a `bowl' is typically {\em on} a `table', a `backpack' is typically {\em behind} a `person', etc. Therefore region proposals that have the characteristic spatial relation of a target class to other source classes should be more probable of belonging to that target class.

To obtain the spatial-relation common-sense matrix $Y_{sp}$, we utilize the information about relative locations and sizes of {\em source} object classes $C_{source}$ in the Visual Genome~\cite{krishna-ijcv2017} knowledge base that contains visually-grounded triplets \{object1, relation, object2\}.

\begin{figure}[t]
    \centering
    \includegraphics[width=0.7\textwidth]{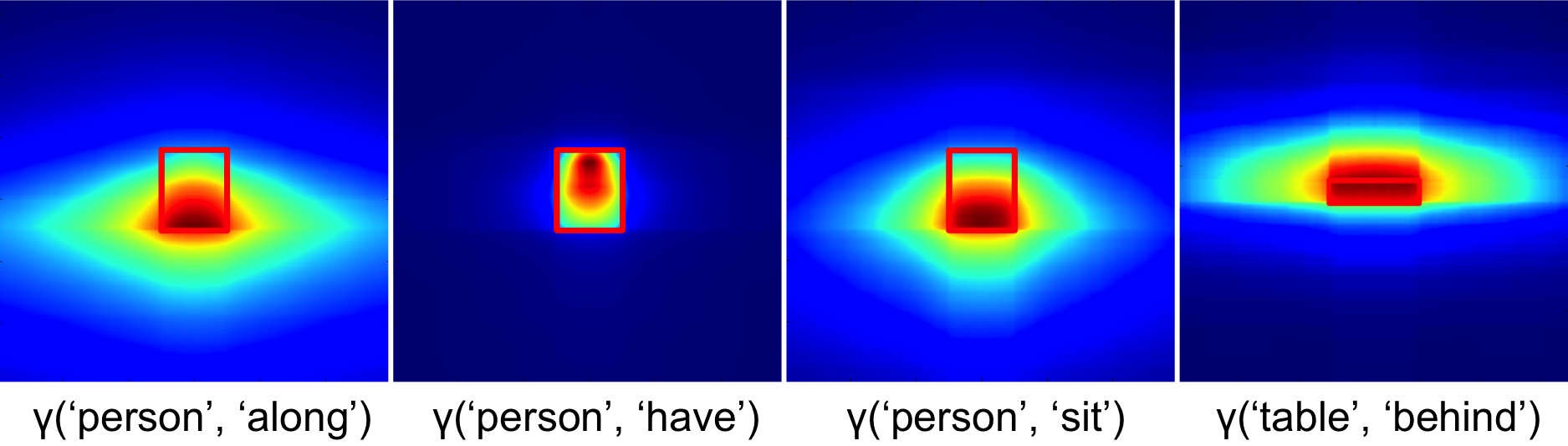}
    \vspace{-0.2in}
    \caption{\textbf{Spatial common-sense distribution:} We model the relative location distributions $\gamma^L_{c_j, rel}$ encoding the pixel-wise probability of all other source objects for a given object $c_j$ (shown as a red box) under a given relation $rel$.}
    \label{fig:spatialmap}
\end{figure}

For each class $c_j$ in $C_{source}$, we model the relative location distributions $\gamma^L_{c_j, rel}$ encoding the pixel-wise probability of all other source objects under a given relation $rel$. For example, Fig.~\ref{fig:spatialmap} shows the distribution of objects with respect to the `along' relationship for the source `person' object class. In a similar way, for each class $c_j$ in $C_{source}$ and a given relation $rel$, we also model the relative size distributions $\gamma^S_{c_j, rel}$. Note that these distributions need to be learned just once using \emph{only the source classes} and then can be reused for any of the target classes (i.e., without bounding box annotations for $C_{target}$ classes).

For a new target class $c_i$ from $C_{target}$, we first gather its most common relation with respect to source classes $c_j$ in $C_{source}$ using the \{object1, relation, object2\} triplet information from~\cite{krishna-ijcv2017}\footnote{For any target class, we gather this information from existing knowledge bases~\cite{krishna-ijcv2017,tandon-wsdm2014} by analyzing the $rel$ that is commonly used to associate it with the source classes.} and then compute the $Y_{sp}$ matrix as:

$$ Y_{sp}(r_n,c_i) =  \max_{c_j \in C_{vis}} \frac{1}{2}(\gamma^L_{c_j,rel}({\bf x}^{center}_{r_n}) + \gamma^S_{c_j,rel}(area_{r_n})),$$

where ${\bf x}^{center}$ and $area$ denote the center coordinate and the size of a proposal, and $C_{vis}$ are the subset of classes from $C_{source}$ that are visible in the given image with their locations determined by running the pre-trained detectors $\phi$.

\begin{figure}[t]
	\centering
	\includegraphics[width=0.95\textwidth]{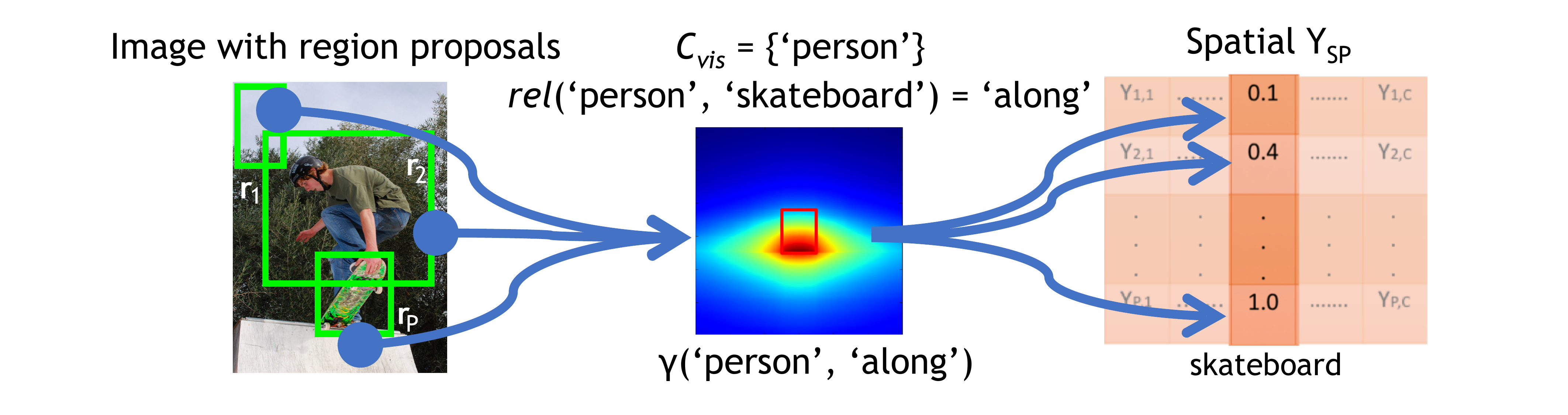}
	\vspace{-0.2in}
	\caption{\textbf{Spatial Common-sense:} The $Y_{sp}(.,skateboard)$ values are computed by measuring the spatial relationship of proposals with respect to the source `person' class. Based on our knowledge base~\cite{krishna-ijcv2017}, `along' is the most common relation between the `skateboard' and `person' class. Observe that the proposal $r_P$ gets the highest $Y_{sp}$ value.}
	\label{fig:commonsense_sp}
\end{figure}

In Fig.~\ref{fig:commonsense_sp}, the proposal $r_P$ gets a higher $Y_{sp}$ value for the `skateboard' class as it is in sync with the `along' relation under the $\gamma^L_{person, along}$ distribution.

\paragraph{Scene Common-sense.} Some objects appear more in certain scenes than others; e.g., a `surfboard' is more likely to be found on a {\em beach}. 
 Hence, images depicting scenes associated with class $c_i$ are more likely to contain instances of $c_i$.
 
To obtain the scene common-sense matrix $Y_{sc}$, we leverage the SceneUNderstanding (SUN)~\cite{xiao-ijcv2016} and Places~\cite{zhou-nips2014} knowledge bases. These databases not only contain information about commonly occurring scene labels for the different object classes but also provide access to pre-trained scene classifiers ($\beta$). Let $SC_{c_i}$ denote the set of scene labels associated with a new target object class $c_i$, and $I$ be the given image containing the proposals, then $Y_{sc}$ can be computed as: $Y_{sc}(r_n,c_i) = \sum_{s_j \in SC_{c_i} }\beta_{s_j}(I), n\in(1,P).$
All proposals $r_n$ in image $I$ get high prior for class $c_i$ if $I$ depicts the scene that frequently contains instances of class $c_i$. Note that this scene common-sense knowledge would be helpful in case of noisy image-level labels (e.g., in the webly-supervised setting), and may not be relevant when we already have clean human-annotated labels that indicate the presence/absence of objects in images.

\section{Experimental Results \& Analysis}

In this section, we report quantitative and qualitative analyses for validating the benefit of transferring common-sense knowledge for object detection. We also conduct ablation studies to dissect the various components of our approach.

\paragraph{Dataset.} Recent state-of-the-art transfer learning approaches for object detection~\cite{hoffman-nips2014,tang-cvpr2016} have limited their analysis to ImageNet~\cite{Russakovsky-IJCV2015}. While this dataset has enabled substantial progress in this domain, we believe the time is ripe for advancing the transfer-learning paradigm to the next level, i.e., the more complex MS COCO~\cite{lin-eccv2014} benchmark. The MS COCO dataset is challenging not only in terms of its diversity (non-iconic views, varying sizes, etc.)  but also in number of classes, and thus poses a tough challenge for methods that do not have access to bounding box annotations for the target classes. We believe our idea of leveraging multiple common-sense would be more relevant to address this challenge.

Recall that, in our framework, we use the PASCAL VOC object detectors as one of our sources of common-sense knowledge (i.e., for similarity and spatial). Hence, we avoid using the 20 VOC classes ($C_{source}$) within the MS COCO dataset and focus our analysis on the remaining 60 object classes ($C_{target}$). We train our network with MS COCO 2015 training images, evaluate on the validation images, and use the standard performance metrics (mAP with 50\% IoU threshold).

\paragraph{Implementation details.} Our base network of~\cite{bilen-cvpr2014} is initialized with VGG-CNN-F~\cite{chatfield-bmvc2014}. We train the network with a learning rate of $10^{-5}$ for the first 20 epochs and $10^{-6}$ for the next 10 epochs. During training, the images are randomly flipped horizontally and scaled to one of 5 scales: $800 \times 608, 656 \times 496, 544 \times 400, 960 \times 720,$ and $1152 \times 864$. During testing, we average the corresponding 10 detection scores and apply NMS with an overlap threshold of 0.4 after filtering out proposals with probability less than $10^{-4}$.  We use MCG~\cite{arbel-cvpr2014} object proposals. In order to combine the common-sense matrices ($Y_a$,$Y_{sp}$,$Y_{s}$), we take the average of the three matrices and obtain $Y_{all}$, which is used to train the base network. Although our approach involves a few task-specific insights to effectively leverage common-sense knowledge, we found our approach to be robust to a wide range of choices when constructing these common-sense matrices (e.g.~number of similar known classes for similarity common-sense, number of common relations between known/unknown classes for spatial common-sense, etc). 
For similarity common-sense, we represent each class name using a 300-d word2vec representation (obtained using pre-trained model on Google news dataset~\cite{mikolov-nips2013}).

\subsection{Quantitative Results}

Table~\ref{table:detection_results} presents the results obtained using our approach and compares it to other relevant approaches. As an upper-bound, we also include the fully-supervised Fast-RCNN~\cite{girshick-iccv2015} detection result (using VGG-CNN-F and trained with bounding box annotations for the target classes) obtaining 25.2\% mAP. While our approach falls well short of this upper bound, it reveals to us the possibility of using common-sense for bridging the gap to fully-supervised methods. \\

\begin{table}[t!]
    \begin{center}
        \begin{tabular}{| l | c | c | c | c |}
            \hline
            \ \ \ \ \ \ Methods & AP & $AP^S$  & $AP^M$ & $AP^L$   \\
            \hline
            Classification Network (`No Adapt'~\cite{hoffman-nips2014}) & 3.4 & 0.9 & 2.9  & 6.1  \\
            LSDA~\cite{hoffman-nips2014} & 4.6  & 1.2 & 5.1 & 7.8 \\
            LSDA+Semantic~\cite{tang-cvpr2016} & 4.7  & 1.1 & 5.1 & 8.0 \\
            LSDA+MIL~\cite{hoffman-cvpr2015} & 5.9  & 1.5 & 8.3 & 10.7 \\
            Fine-tuned Detection Network & 10.8 & 1.2 &8.9  & 18.6  \\
            \hline
            Proposed Approach    & \bf{14.4} & \bf{2.0} & \bf{12.8} & \bf{24.9}   \\
            \hline \hline
            Oracle: Full Detection Network~\cite{girshick-iccv2015}    & 25.2 & 5.8 & 26.0 & 41.6  \\
            \hline
        \end{tabular}
        \caption{Detection results on MS COCO validation set. Our proposed approach substantially improves over existing transfer-learning methods.}
        \label{table:detection_results}
    \end{center}
    \vspace{-0.5in}
\end{table}

\noindent{\emph{\bf Comparison to Transfer-learning.}} The most relevant state-of-the-art transfer learning methods are LSDA~\cite{hoffman-nips2014},~\cite{hoffman-cvpr2015} and~\cite{tang-cvpr2016}.  However, as~\cite{hoffman-nips2014,hoffman-cvpr2015,tang-cvpr2016} neither report results on the challenging MS COCO dataset nor provide their training code, we re-implemented LSDA~\cite{hoffman-nips2014}\footnote{Similar setting as ours: use image-level labels for all training images of MS COCO and VOC, use bounding box for only 20 VOC classes, and VGG-CNN-F as base network.}. Running this baseline on the 60 MS COCO $C_{target}$  classes yields 4.6\% mAP, which is substantially inferior than our proposed approach (14.4\%). We hypothesize the poor performance of the LSDA frameworks on the MS COCO dataset to the following reasons:

(i) LSDA approaches~\cite{hoffman-nips2014,hoffman-cvpr2015,tang-cvpr2016} are inherently limited to exploiting only {\em similarity} common-sense. While similarity suffices in case of object-centric datasets, more richer common-sense (such as {\em attribute}, {\em spatial}, etc., that we use) needs to be leveraged when dealing with complex scene-centric datasets. Further when the size of $C_{source}$ is small (e.g., set of 20 classes in our MS COCO experiments), similarity fails to work well in gathering meaningful neighbors between $C_{target}$ and $C_{source}$ classes. As a result, LSDA methods cannot transform the classifier successfully. Particularly, the class-invariant transformation of the weights from conv1 to fc7 layers in LSDA will not generalize well when the similarity overlap between the classes in $C_{source}$ and $C_{target}$ is poor. Our approach alleviates this concern by explicitly using detection probabilities of the $C_{source}$ classes for each image (rather than doing a weight transformation), and also by using other common-sense knowledge ({\em attribute}, {\em spatial}, etc., which would weigh more in the absence of similar $C_{source}$ classes).

(ii) As MS COCO has images with multiple categories and small objects, the  initial classifier network learned in the LSDA methods is poor (giving only $3.4\%$ mAP, see Table~\ref{table:detection_results} top-row) and therefore transforming it results in a poor detector. Our approach alleviates this concern by employing a proposal classification and ranking detection network (11.7\% mAP, see Table~\ref{table:ablation}, `Base network').

(iii) Finally, class co-occurrences in images poses a formidable challenge for the LSDA methods. For example, in case of the target class `spoon', the top nearest neighbor (among the 20 $C_{source}$ classes) is `dining-table', which often co-occurs with it and thus transferring its detection knowledge will cause confusion to the detector. Our approach is robust to such cases of incorrect similarities as the similarity information is only used as a `prior' during our base detection network training.

\noindent \emph{Transfer learning by fine-tuned detection network.} We also explored an alternate transfer-learning strategy, where we initialize our base detection network (Section~\ref{sec:baseline}) with pre-trained Fast-RCNN weights from the source categories and then fine-tune it on the target classes using only image-level labels. While this method produces a relatively higher mAP of 10.8\% than LSDA, it is still lower than our approach. We believe this is due to the network weights getting overfit to the 20 $C_{source}$ classes and subsequently failing to generalize well to the 60 $C_{target}$ classes. In contrast, our approach does not have this overfitting issue as we initialize our base detection network with the weights from a more general network (trained on 1000 ImageNet classes) and then use the 20 $C_{source}$ classes pre-trained detection models only for computing similarity common-sense. \\

\noindent{\emph{\bf Alternate approaches for leveraging common-sense.}}  To analyze the significance of our proposed approach for leveraging common-sense, we also studied alternative strategies for leveraging common sense cues that use exactly the same external knowledge/bounding box information as our approach.

The first strategy uses common-sense as a contextual post-processing tool~\cite{divvala-cvpr2009}.  Specifically, for a test image, we compute the common-sense matrix $Y$, and then modulate its classification matrix $X$ via element-wise multiplication to produce the final score matrix: $Z_{test} = X_{test}\cdot Y_{test}$. Table~\ref{table:ablation} (`Post-process') displays the result obtained using this post-processing strategy, which obtains $11.8\%$ (and $14.1\%$ when common-sense matrix $Y$ is also used during training). Observe that the post-processing result is lower compared to our approach, which transfers common-sense during training only and not during testing. This indicates that $X$ has already incorporated the common-sense knowledge during training, and thus using $Y$ is redundant at test time. When $Y$ is introduced during testing, it is difficult for common sense to fix any incorrect biases that the detector may have learned (e.g., focusing only on the most discriminative part of an object). It may even hurt as any mistakes in common-sense information $Y$ cannot be circumvented when used directly at test time. In contrast, by transferring common sense during training, it can guide the algorithm to \emph{learn} to focus on the correct image regions.

The second alternate strategy for leveraging common-sense analyzes the possibility of improving detection performance by simply having access to pre-trained object/spatial/attribute classifiers.  In this case, we trivially append a 45-dimensional feature to the fc7 appearance feature of each proposal during training. The first 20 dimensions correspond to the detection probabilities of the 20 $C_{source}$ classes and the remaining 25 dimensions correspond to the attribute probabilities of the classifiers ($\theta$) pre-trained on the ImageNet attribute knowledge base. While this model yields a boost of 1.0\% mAP compared to the base detection network (see Table~\ref{table:ablation} `Feature'), it is 1.7\% lower than our proposed model. This reveals that merely concatenating external knowledge features to visual appearance features is insufficient.

\begin{table}[t!]
    \begin{center}
        \begin{tabular}{| c | c || c | c || c | c | c | c || c |}
            \hline
            \multirow{2}{*}{\emph{Method}} &
            \multirow{2}{*}{\emph{Base network}} &
            \multicolumn{2}{|c||}{\emph{Alternatives}} &
            \multicolumn{4}{|c||}{\emph{Ablations}} &
            \multicolumn{1}{|c|}{\emph{Ours}} \\ \cline{3-9}
            && Post-process & Feature & Attr & Spatial & Sim & Joint & +Sim(Bbox)   \\
            \hline
            \emph{mAP} & 11.7 & 11.8/14.1 & 12.7 & 12.2 & 13.0 & 13.7 & 14.1 & \textbf{14.4} \\
            \hline
        \end{tabular}
        \caption{Analysis on the MS COCO dataset: `Base network' is the result obtained by using our base detection network using no common-sense. `Alternatives' are the alternate strategies for leveraging common-sense knowledge. Their performance is lower than our model demonstrating the benefit of our proposed approach. `Ablations' show the improvement obtained using each common-sense cue over the base network. Combining all common-sense cues produces the best result indicating their complementarity.  }
        \label{table:ablation}
    \end{center}
    \vspace{-0.5in}
\end{table}

\subsection{Ablation \& Qualitative Analysis}
We also analyzed the importance of the various common-sense cues in our proposed approach. Table~\ref{table:ablation} `Ablations' shows that each common sense gives a boost over the base network, which only relies on appearance cues. Among the individual cues, we see that similarity helps the most. (Scene cue was not explored in this scenario and its influence will be analyzed in the webly-supervised setting.) Combining attribute, spatial, and similarity common sense (`Joint') leads to a greater boost of 2.4\% mAP, which shows their complementarity. Finally, we also borrow the bounding box regressors trained on the similar $C_{source}$ classes and apply them to the $C_{target}$ classes, which further boosts performance to 14.4\%.

Taking a closer look at individual classes, we find that using the attribute common-sense for `oven' (that is usually {\em white/black}) results in a boost of 7.3\% AP. By using the spatial common-sense for `frisbee' with respect to the source object `person', we get an improvement of 10.5\% in AP, whereas using the spatial relation for a `bowl' with respect to the source object `table' gives a boost of 2.2\%.  For `giraffe' and `bed', we use the common-sense that they are semantically similar to \{`sheep', `horse', `dog', `cow', `bird', `cat'\}, and `couch' respectively, which leads to an improvement of 28.7\% and 12.1\% AP, respectively.

We next analyze the importance of using word2vec for similarity common-sense. For this, we replace word2vec similarity with visual similarity, which results in $12.1\%$ compared to our $13.7\%$. Word2vec similarity is more robust than visual similarity (used in LSDA~\cite{hoffman-nips2014}), particularly for challenging dataset like MS COCO (with small objects and co-occurring objects from different classes). We also tried WordNet~\cite{miller-acm1995} treepath based similarity which also gives an inferior result of $13.1\%$.

\begin{figure*}[t!]
    \centering
    \includegraphics[width=0.93\textwidth]{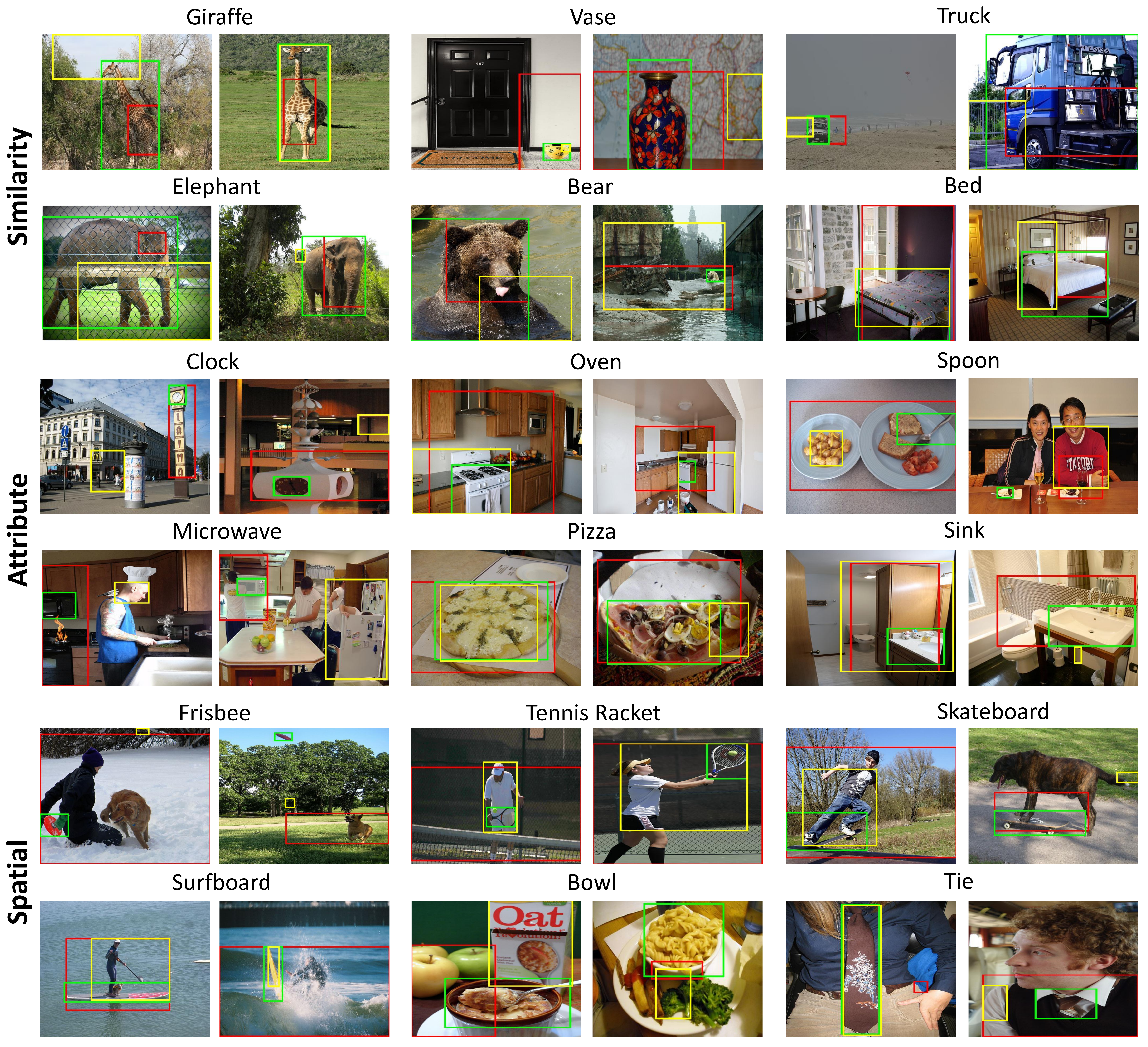}
    \caption{Qualitative detection results on MS COCO (Ours: green boxes; Base network: red; LSDA+Semantic~\cite{tang-cvpr2016}: Yellow):  Observe that our approach produces better detections than the base network for all three common-sense. 
    For `giraffe' and `elephant', by using similarity common-sense (i.e., being similar to other animal categories in $C_{source}$), our approach detects the full body extent rather than localizing a discriminative body part. By using spatial and attribute common-sense, e.g., `clock' being {\em round}, `spoon' being {\em metallic}, and `microwave' being {\em white/black}, we get better detections. 
    }
    \label{fig:qualitative}
\end{figure*}

\begin{figure}[t!]
	\centering
	\includegraphics[width=0.7\textwidth]{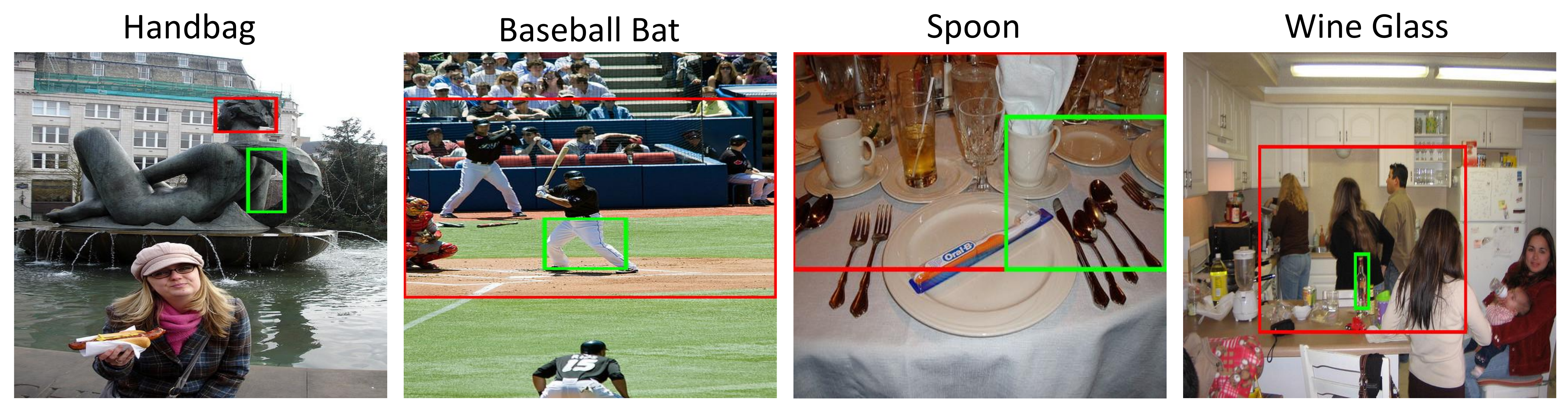}
	\caption{{\small Example failures: Our approach fails when the object-of-interest is hardly-visible (`handbag') or when source objects with similar attribute ({\em metallic}) are cluttered together (`spoon'). For `wine glass', we falsely detect the `bottle' because during training we provided the common-sense that wine-glass is semantically similar to a bottle.}}
	\label{fig:failure}
\end{figure}

Fig.~\ref{fig:qualitative} shows some qualitative detections (for each of the common-sense) produced by our approach (green box) and compares them to competing baselines. We can observe that using common-sense helps improve performance. For example, by using spatial common-sense, our approach gets rid of the co-occurring background for `frisbee' and `surfboard' (person and water, respectively). Interestingly, for `frisbee', our approach uses the spatial common-sense that a `frisbee' often spatially-overlaps with a `person' in order to learn its appearance during training.  It is then able to detect a `frisbee' at test time even when a `person' is not present (see `frisbee' image in the second column). This indicates that while our approach leverages common-sense as a prior during training to learn about the object's appearance, at the same time, our network is not dependent on common-sense during testing. 
Fig.~\ref{fig:failure} shows some failure cases.

\paragraph{Towards webly-supervised detection using common-sense supervision.} What happens when we apply our method in cases when we do not even have explicit human-annotated image labels for the target class? This is exactly the setting studied in the webly-supervised domain where images retrieved from the web are used for training detection models.

We conducted a preliminary investigation wherein we ran our proposed approach on training images retrieved from the web (i.e., instead of the MS COCO training imageset). As images retrieved from web are potentially noisy, common-sense knowledge should be particularly useful in mitigating the noise. Our preliminary results indicate that our proposed idea is promising even in the webly setting (Base network~\cite{bilen-cvpr2016}: 6.8\%, vs. Ours: 8.3\%).

Further, to analyze potential concerns about the generalizability of our acquired common-sense knowledge, we also tested these webly+commonsense models on the ImageNet 200 detection valset~\cite{Russakovsky-IJCV2015}. Even in this case, our approach yields interesting performance gains (Base network~\cite{bilen-cvpr2016}: 6.2\%, vs. Ours: 8.8\%).

\vspace{-0.1in}
\section{Conclusion}
\vspace{-0.1in}
In this paper, we presented DOCK, a novel approach for transferring common-sense knowledge from a set of categories with bounding box annotations to a set of categories that only have image-level annotations for object detection. We explored how different common-sense cues based on similarity, attributes, spatial relations, and scene could be jointly used to guide the algorithm towards improved object localization. Our experiments showed that common-sense knowledge can improve detection performance on the challenging MS COCO dataset. We hope our work will spur further exciting research in this domain.

\paragraph{\textbf{Acknowledgements.}}
\vspace{-0.1in}
This work is in part supported by ONR N00014-13-1-0720, NSF
IIS-1338054, NSF IIS-1748387, NSF IIS-1751206, NSF-1652052, NRI-1637479, ARO YIP W911NF-17-1-0410, Allen Distinguished
Investigator Award, Allen Institute for AI, Microsoft Azure Research Award and the GPUs donated by NVIDIA. A part of this work was done while Krishna Kumar Singh was an intern at Allen Institute for AI.

\bibliographystyle{splncs04}
\bibliography{0615}
\end{document}